\documentclass[letterpaper, 10 pt, conference]{ieeeconf}  

\IEEEoverridecommandlockouts                              

\usepackage{graphics} 
\usepackage{epsfig} 
\usepackage{times} 
\usepackage{amsmath} 
\usepackage{amssymb}  
\usepackage{mathrsfs}
\usepackage{algorithm}
\usepackage{algorithmicx}
\usepackage{algpseudocode}
\usepackage{cite}
\usepackage{booktabs}
\usepackage{epstopdf}
\usepackage{subfigure,color,balance}
\usepackage{verbatim}
\usepackage{cases}
\usepackage{enumerate}
\usepackage{balance}
\usepackage{threeparttable}

\usepackage[colorlinks=true,      
linkcolor=black,      
citecolor=black,      
filecolor=black,      
urlcolor=blue]{hyperref}

\def\0{{\bf 0}}
\def\1{{\bf 1}}

\title{\LARGE \bf
RADE: Learning Risk-Adjustable Driving Environment via \\ Multi-Agent Conditional Diffusion
}

\author{Jiawei Wang$^1$, Xintao Yan$^{1}$, Yao Mu$^{2}$, Haowei Sun$^{1}$, Zhong Cao$^{1}$, and Henry X. Liu$^{1,3*}$
\thanks{This work was partially funded by the United States National Science
Foundation through the Mcity 2.0 Project (CMMI \#2223517). $^*$Corresponding author: Henry X. Liu (henryliu@umich.edu)}
\thanks{$^{1}$J.~Wang, X.~Yan, H.~Sun, Z.~Cao, and H.~X.~Liu are with the Department of Civil and Environmental Engineering, University of Michigan, Ann Arbor, MI 48109, USA.} 
\thanks{$^{2}$Y.~Mu is with the Department of Computer Science, University of Hong Kong, Hong Kong 999077, China. }
\thanks{$^{3}$H.~X.~Liu is also with University of Michigan Transportation Research Institute, Ann Arbor, MI 48109, USA. }
}%

\begin{document}

\maketitle
\thispagestyle{empty}
\pagestyle{empty}

\begin{abstract}
Generating safety-critical scenarios in high-fidelity simulations offers a promising and cost-effective approach for efficient testing of autonomous vehicles. Existing methods typically rely on manipulating a single vehicle's trajectory through sophisticated designed objectives to induce adversarial interactions, often at the cost of realism and scalability. In this work, we propose the Risk-Adjustable Driving Environment (RADE), a simulation framework that generates statistically realistic and risk-adjustable traffic scenes. Built upon a multi-agent diffusion architecture, RADE jointly models the behavior of all agents in the environment and conditions their trajectories on a surrogate risk measure. Unlike traditional adversarial methods, RADE learns risk-conditioned behaviors directly from data, preserving naturalistic multi-agent interactions with controllable risk levels.  To ensure physical plausibility, we incorporate a tokenized dynamics check module that efficiently filters generated trajectories using a motion vocabulary. We validate RADE on the real-world rounD dataset, demonstrating that it preserves statistical realism across varying risk levels and naturally increases the likelihood of safety-critical events as the desired risk level grows up. Our results highlight RADE’s potential as a scalable and realistic tool for AV safety evaluation.


\end{abstract}

\section{Introduction}


Ensuring the safety of autonomous vehicles (AVs) requires rigorous evaluation under safety-critical scenarios, such as near-misses and crashes~\cite{cao2023continuous}. However, collecting such data in the real world is inherently difficult, not only because of its rarity, but also due to ethical and logistical constraints~\cite{liu2024curse}. Alternatively, high-fidelity simulation environments provide a controllable, scalable, and cost-effective approach~\cite{suo2021trafficsim}. These environments enable systematic generation of diverse and safety-critical scenarios that are difficult to observe or replicate in the real world. A key challenge in building such simulations is to generate background vehicle behaviors that not only capture high-risk interactions but also remain naturalistic and consistent with real-world driving patterns~\cite{yan2023learning}.

Current research on safety-critical traffic simulation mostly focuses on single-agent adversarial attacks. Precisely, they seek to manipulate the trajectory of one background vehicle to induce collisions or near-miss events against the AV. Along this direction, early learning-based approaches often employ reinforcement learning (RL) to train highly adversarial policies for background vehicles; see, \emph{e.g.},~\cite{koren2018adaptive,kuutti2020training}. However, these methods typically suffer from poor realism and diversity, as the learned policies are rarely grounded in real-world driving data. More recently, deep generative models have been used to generate adversarial behaviors from real-world datasets. Typically, a two-step framework is adopted: 1) first, learn normal driving behaviors from real-world datasets; and 2) then, perturb an adversary vehicle’s behavior by optimizing towards predefined safety-critical objectives. For example, STRIVE~\cite{rempe2022generating} refines adversarial trajectories by optimizing within the latent space of a Variational Autoencoder (VAE), while DiffScene~\cite{xu2023diffscene}, Safe-Sim~\cite{chang2024safe}, and AdvDiffuser~\cite{xie2024advdiffuser} leverage guided sampling techniques to generate adversarial trajectories through carefully designed optimization formulations in each denoising step of a diffusion model.

However, these perturbation-based adversarial methods introduce inherent trade-offs. While behavior realism is achieved by learning the normal behavior models from data, it is again compromised through the process of injecting adversarial intent, since it directly modifies the behavior model output. Consequently, unrealistic or overly aggressive maneuvers may be obtained that may not align with plausible human driving behaviors. To evaluate realism more systematically, statistical realism has been established as an effective metric, which captures distribution-level accuracy with real-world driving pattern~\cite{yan2023learning,sun2025terasim}. In addition, these frameworks mostly rely on manually predefined activation schemes, where a specific background vehicle is designated to execute a particular adversarial maneuver at a specific time. As a result, these methods are lack of flexibility and scalability due to limitations to localized safety-critical interactions involving one single adversary vehicle, rather than generating an entire environment where multiple vehicles naturally and dynamically interact to create emerging risky behaviors. For reliable and efficient stress-testing of AV systems, there is a significant need for a simulation environment with both statistical realism and naturally generated risk.

\begin{figure}[t]
	\vspace{1mm}
	\centering
	{\includegraphics[width=0.48\textwidth]{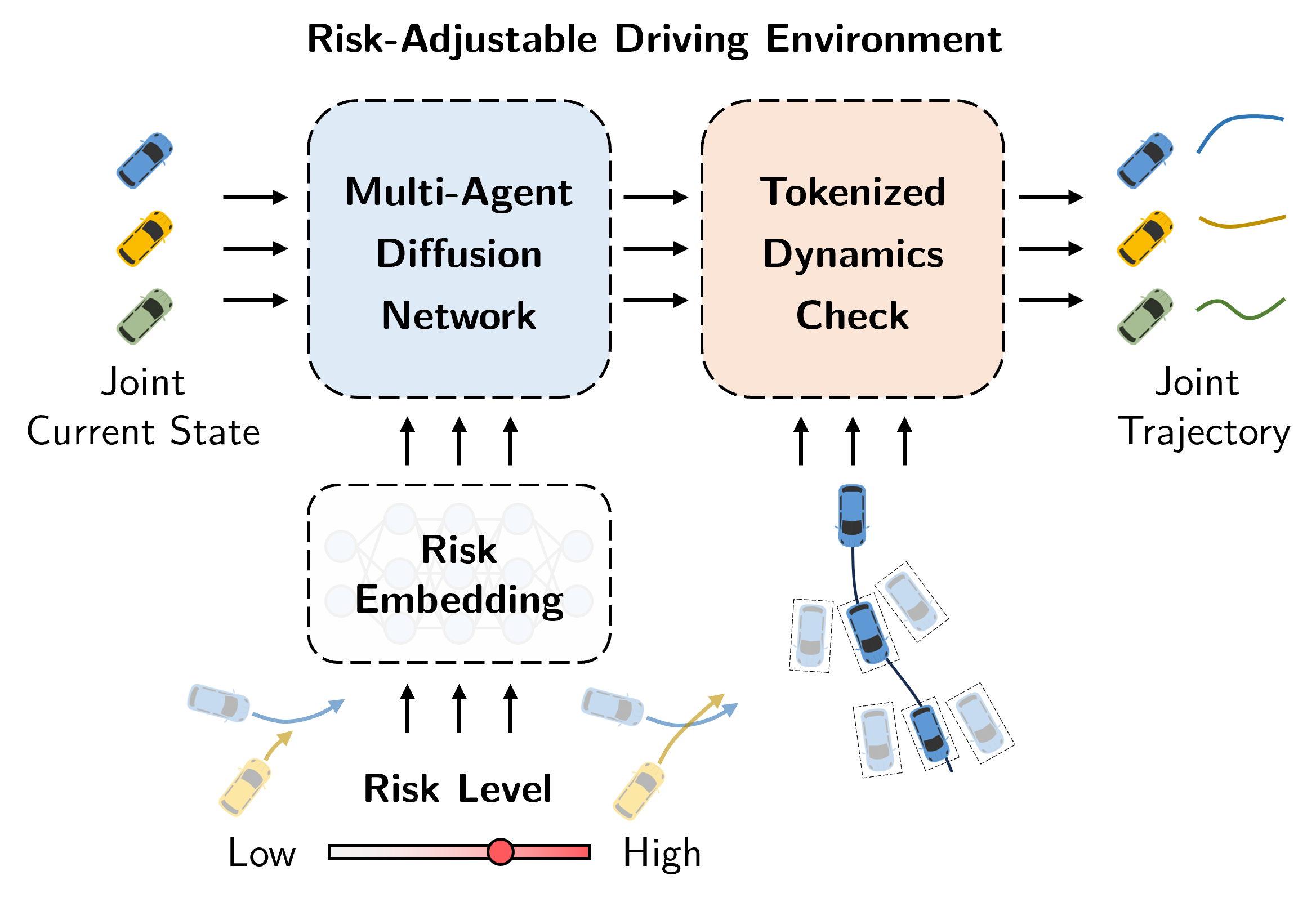}}
	\caption{The proposed RADE framework. Conditioned on the desired risk level, RADE takes joint current state of all the vehicles as the input, and generates raw trajectories using a multi-agent diffusion network, which conducts denoising steps over vehicle states. These trajectories are then refined by a tokenized dynamics check module, producing the final valid joint future trajectories.}
	\label{Fig:Framework}
\end{figure}

To address these limitations, we propose a Risk-Adjustable Driving Environment (RADE), where background vehicles exhibit statistically realistic and risk-adjustable behaviors. To achieve this, we leverage a recently proposed multi-agent diffusion framework~\cite{zhu2024madiff} and introduce risk-conditional trajectory generation; see Fig.~\ref{Fig:Framework} for the system framework. Unlike traditional single agent based adversary generation, our approach models the joint behavior of all agents in the traffic scene and creates a driving environment with complex interactions. By multi-agent modeling, RADE captures the collective emergence of traffic risk. Instead of manipulating agent behaviors by enforcing adversarial interactions, our approach aims to directly learn the trajectory given varying risk levels, which can be defined based on surrogate safety measures widely used in transportation domain. Specifically, a Post-Encroachment Time (PET)~\cite{peesapati2018can} related criterion is designed in this paper.  Our risk conditioning approach ensures that the generated traffic environment remains statistically realistic, while also enabling efficient control over risk levels in simulations. Besides, to ensure physical plausibility, a tokenized dynamics check module is designed to filter the diffusion generated trajectories in an efficient way. 
Through extensive random experiments on a roundabout scenario, we demonstrate that RADE preserves statistical realism across a range of risk levels. Moreover, the crash rate in the simulated environment increases consistently with higher desired risk values, confirming that RADE can effectively generate realistic yet risk-adjustable driving scenarios.


\section{Related Work}

\subsection{Behavior Modeling for Traffic Simulation}

Early traffic simulations mostly employ rule-based methods~\cite{treiber2000congested,kesting2007general} to model individuals' driving behaviors, which fail to achieve human-like behavior realism despite its ability to capture traffic flow dynamics. Recently, data-driven methods have received increasing attention for traffic simulations with realistic and complex interactions~\cite{yan2023learning}. BITS~\cite{xu2023bits} introduces a bi-level imitation learning framework by decoupling driving behaviors into high-level intent inference and low-level driving control. TrafficBots~\cite{zhang2023trafficbots} presents configurable behavior generation based on goal conditioned VAE. SMART~\cite{wu2024smart} proposes a tokenized motion generation framework using decoder-only Transformer. The aforementioned research focuses on normal behavior modeling. For safety-critical behavior generation, existing simulation work mostly seeks to design an adversary trajectory for one single background vehicle, and can be divided into two category: 1) training an adversary policy for the background vehicle via RL~\cite{koren2018adaptive,kuutti2020training,feng2023dense}; or 2) perturbing the normal driving behavior to generate safety-critical results~\cite{hanselmann2022king,zhang2023cat}. Both categories can lead to unrealistic or over-aggressive maneuvers, which limits their practical applicability to efficient AV stress-testing.

\subsection{Diffusion Models for Traffic Simulation}

Diffusion Models have shown significant promise in generative tasks~\cite{ho2020denoising}. Particularly, conditional diffusion models have the great potential for controllable trajectory generation from logged data~\cite{janner2022planning,ajay2022conditional}, which can be naturally applied to controllable and realistic traffic simulations. CTG~\cite{zhong2023guided} designs rule-based functions to guide the denoising network output for traffic rule compliance, with CTG++~\cite{zhong2023language} further utilizing Large Language Models for translating user queries into differentiate guidance functions. VBD~\cite{huang2024versatile} and SceneDiffuser~\cite{jiang2024scenediffuser} incorporates behavior prediction priors into the diffusion models for enhanced closed-loop realism. The aforementioned methods mostly employ the guided sampling technique for reasonable and controllable behavior generation, which modifies the model output by predefined optimization formulations. By similar means yet with adversarial objectives in the optimization, the diffusion model can be guided to generate safety-critical behaviors for the adversary vehicle. Most diffusion based research follows this direction to control one adversary vehicle and challenge the AV; see, \emph{e.g.},~\cite{chang2024safe,xu2023diffscene}. However, realism is not guaranteed through this guidance process, and existing work is limited to generating single vehicle adversaries rather than building a risk-adjustable driving environment.

\section{Problem Statement}
\label{Sec:3}


In this section, we first introduce the basics of diffusion models and then present the problem statement. 

\subsection{Diffusion Models}

To generate data samples that follow the data distribution $q(\boldsymbol{\tau})$ from a dataset $\mathcal{D}$, diffusion models adopt a procedure of reversing a forward noising process. The noising process is normally defined as
\begin{equation} \label{Eq:noising}
    q(\boldsymbol{\tau}^k | \boldsymbol{\tau}^{k-1}) = \mathcal{N}(\boldsymbol{\tau}^k;\sqrt{1 - \beta_k} \boldsymbol{\tau}^{k-1}, \beta_k \boldsymbol{I}),
\end{equation}
which starts from an original data sample $\boldsymbol{\tau}^0$ from the dataset and provides a sequence of increasingly noisy trajectories $\boldsymbol{\tau}^1,\ldots,\boldsymbol{\tau}^K$. With predefined variance schedule $\beta_k$ and long enough $K$, the final noisy sample $\boldsymbol{\tau}^K$ approaches an isotropic Gaussian distribution $\mathcal{N}(\boldsymbol{0},\boldsymbol{I})$. Then the noising process is reversed by learning a denoising process
\begin{equation} \label{Eq:denoising}
p_\theta(\boldsymbol{\tau}^{k-1} | \boldsymbol{\tau}^k) = \mathcal{N}(\boldsymbol{\tau}^{k-1};\boldsymbol{\mu}_\theta(\boldsymbol{\boldsymbol{\tau}}^k,k), \boldsymbol{\Sigma}_k),
\end{equation}
where $\boldsymbol{\Sigma}_k$ is a fixed variance schedule. Starting from Gaussian noise $\boldsymbol{\tau}^K\sim\mathcal{N}(\boldsymbol{0},\boldsymbol{I})$, iteratively running~\eqref{Eq:denoising} provides a series of denoising samples until $\boldsymbol{\tau}^0$. As revealed in~\cite{ho2020denoising}, the estimated mean value $\boldsymbol{\mu}_\theta(\boldsymbol{\boldsymbol{\tau}}^k,k)$ can be calculated from an estimated noise $\epsilon_{\theta}(\boldsymbol{\tau}^k, k)$ by a closed form. Thus, the reverse process can be trained on a simplified surrogate loss:
\begin{equation} \label{Eq:noise}
    \mathcal{L}(\theta) = \mathbb{E}_{k\sim[1,K],\boldsymbol{\tau}^0 \in \mathcal{D}, \epsilon\sim\mathcal{N}(\boldsymbol{0},\boldsymbol{I})} \left\| \epsilon - \epsilon_{\theta}(\boldsymbol{\tau}^k, k) \right\|^2,
\end{equation}
which estimates the noise $\epsilon\sim \mathcal{N}(\boldsymbol{0},\boldsymbol{I})$ that is added to the clean sample $\boldsymbol{\tau}^0$ to generate the noisy sample $\boldsymbol{\tau}^k$.

\subsection{Problem Statement}

Given a traffic scenario, like roundabout, we denote $N$ as maximum vehicle number, $T$ as episode time length, and $D$ as the state dimension of each vehicle at each time step. Denote $\boldsymbol{\tau}_i =  [\boldsymbol{s}^{(0)}_i,\boldsymbol{s}^{(1)}_i,\ldots,\boldsymbol{s}^{(T)}_i ]$ as the trajectory of vehicle $i$. Then, the joint trajectories of all the vehicles are represented as the following tensor:
\begin{equation}
    \boldsymbol{\tau} = [
\boldsymbol{\tau}_1;\boldsymbol{\tau}_2;\ldots;\boldsymbol{\tau}_N
    ] \in \mathbb{R}^{N\times T \times D}.
\end{equation}

Our main objective is to generate a realistic driving environment given a desired risk level $r$  ($0 \leq r \leq 1$) for the entire traffic scenario. Particularly, when the desired risk level grows up, the simulation environment should have a higher likelihood for safety-critical events, while statistical realism is preserved. To achieve this, the following conditional generative problem can be formulated:
\begin{equation}
    \max_\theta \mathbb{E}_{\boldsymbol{\tau}\sim\mathcal{D}} \log p_\theta(\boldsymbol{\tau} | r),
\end{equation}
which learns a $p_\theta$ that estimates the conditional distribution of joint trajectory $\boldsymbol{\tau}$ given the desired risk $r$. Then, the joint trajectory generation model 
is run in an auto-regressive manner for roll-out simulations. 

Note that unlike most existing work~\cite{zhong2023guided,chang2024safe,xu2023diffscene}, which independently generates ego vehicle's trajectory by conditioning on background vehicles' information, we aim to generate the joint trajectories of all the vehicles via multi-agent diffusion. This could capture complex interactions in a more natural way~\cite{zhu2024madiff}. In this study, the current state of all the vehicles is treated as the context information for the model input.

\section{Methodology}

In this section, we introduce the multi-agent conditional diffusion framework for building the risk-adjustable driving environment.

\subsection{Conditional Diffusion}

Instead of diffusion over action only~\cite{zhong2023guided} or state-action~\cite{janner2022planning} sequences, in this paper we choose to diffuse over state only trajectories, which is claimed to promise better performance due to the less smooth property of actions~\cite{ajay2022conditional}. Motivated by~\cite{yan2023learning}, we choose each vehicle's state as $\boldsymbol{s}=[p_x,p_y,\cos\theta,\sin\theta]$, where $p_x,p_y$ denote the vehicle coordinates and $\theta$ denotes the vehicle heading.

To generate the traffic scenario given a desired risk $r$, we need to train a diffusion model conditioned on the risk $r$, which is labeled in the offline dataset. To address this, we follow the classifier-free guided generation procedure~\cite{ho2022classifier}, detailed as follows.

\vspace{0.2em}
\noindent\textbf{Sampling Procedure.} The specific sampling process is as follows: Starting from $\boldsymbol{\tau}^0\sim \mathcal{N}(0,\alpha\boldsymbol{I})$, where $0<\alpha\leq 1$ denotes the low-temperature scaling factor, the conditional diffusion process follows the form of:
\begin{equation} \label{Eq:denoising_with_condition}
p_\theta(\boldsymbol{\tau}^{k-1} | \boldsymbol{\tau}^k, r) = \mathcal{N}(\boldsymbol{\tau}^{k-1};\boldsymbol{\mu}_\theta(\boldsymbol{\boldsymbol{\tau}}^k,r,k), \boldsymbol{\Sigma}_k).
\end{equation}
For extracting the distinct portions of trajectories from the dataset with the risk $r(\boldsymbol{\tau})$, a perturbed noise is applied at each denoising step:
\begin{equation}
    \hat{\epsilon} := \epsilon_{\theta}(\boldsymbol{\tau}^k, \varnothing, k) + \omega\left(\epsilon_{\theta}(\boldsymbol{\tau}^k, r(\boldsymbol{\tau}^k), k) - \epsilon_{\theta}(\boldsymbol{\tau}^k, \varnothing, k)\right),
\end{equation}
where $\epsilon_{\theta}(\boldsymbol{\tau}^k, \varnothing, k)$ denotes the unconditional noise, and $\omega$ denotes the condition guidance scale.
Note that the first state of the denoised trajectory is fixed to the current actual state at each denoising step for trajectory validity~\cite{janner2022planning}.

\vspace{0.2em}
\noindent\textbf{Training.} Given a traffic scenario dataset $\mathcal{D}$ containing the joint trajectory of all the vehicles, we train the conditional diffusion process $p_\theta$ via the predicted noise model $\epsilon_\theta$. The loss function is defined as:
\begin{equation}
\mathcal{L}(\theta) = \mathbb{E}_{k,\boldsymbol{\tau}^0, \epsilon,\beta} \left\| \epsilon - \epsilon_{\theta}\left(\boldsymbol{\tau}^k, (1-\beta)r(\boldsymbol{\tau}^0)+\beta \varnothing, k\right) \right\|^2,  
\end{equation}
where $\beta \in \mathrm{Bern}(p)$ is sampled from a Bernoulli distribution. This means that the conditioning value $r$ is ignored with a probability $p$  for balance between training the conditional or unconditional models. During the training process, given original data $\boldsymbol{\tau}^0 \in \mathcal{D}$, step $k \sim \mathcal{U}\{1,\ldots,K\}$ and noise $\epsilon \in \mathcal{N}(\boldsymbol{0},\boldsymbol{I})$, we calculate the risk value for the clean sample $r(\boldsymbol{\tau}^0)$, which will be detailed in Section~\ref{Sec:Risk}, and get the noisy sample $\boldsymbol{\tau}^k$ following the procedure in~\eqref{Eq:noising}. Then we predict the noise $\epsilon_\theta$. Note that similarly to the sampling procedure, the first state of the noisy sample $\boldsymbol{\tau}^k$ is also fixed to the current actual state.

\subsection{Multi-Agent Diffusion Network}

\begin{figure}[t]
	\vspace{1mm}
	\centering
	\subfigure[]
	{\includegraphics[width=0.45\textwidth]{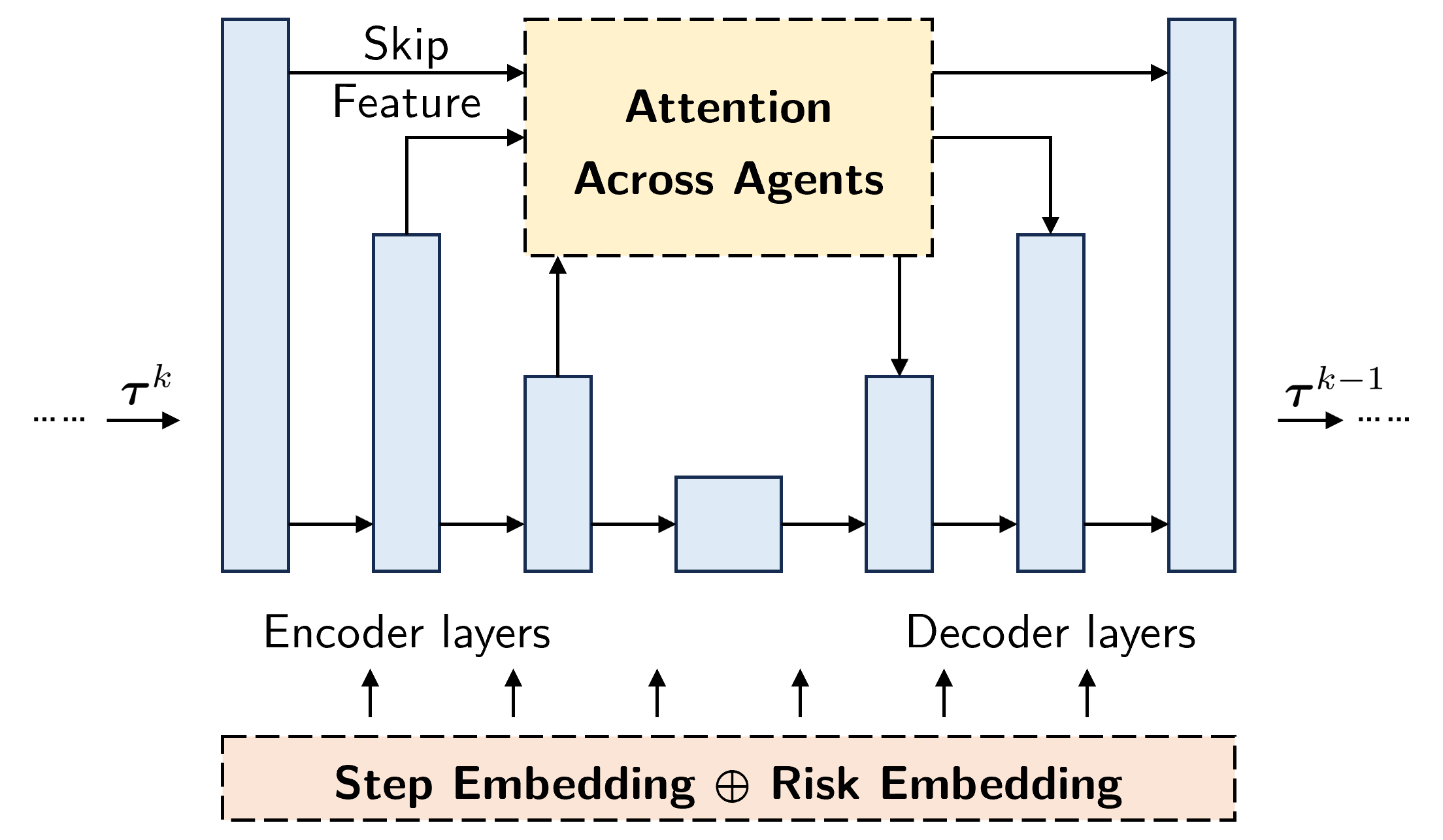}}
	\caption{Network architecture based on the MADiff backbone~\cite{zhu2024madiff}. The model adopts a U-Net structure with multi-agent attention applied to skip-connected features, and receives a risk latent vector at each layer.}
	\label{Fig:Network}
\end{figure}

In this work, we adopt a very recent MADiff architecture~\cite{zhu2024madiff} as the backbone for our learning task, leveraging its attention-based multi-agent diffusion property to model interactive vehicle behaviors. As shown in Fig.~\ref{Fig:Network}, the network is built on a U-Net structure, which is an encoder-decoder architecture with skip connections and can be utilized to model single agent's trajectory. Each encoder or decoder is composed of repeated one-dimensional convolutional residual blocks, followed by down-sampling or up-sampling operations. During decoding, the inputs consist of two parts: a skip-connected feature from the corresponding encoder layer and an embedding from the previous decoder layer. For multi-agent interaction, the distinction lies in the skip connections. Precisely, MADiff applies multi-head attention operations to the skip-connected features, allowing cross-agent feature fusion at multiple hierarchical levels. This setup ensures that each agent’s behavior is influenced by its surroundings, improving realism and coordination in trajectory generation. Note that the risk value $r$ is embedded as a $128$-dimensional latent vector using a multi-layer perceptron (MLP). The risk embedding is then concatenated together with the diffusion step embedding for $k$ and fed into each encoder and decoder layer in the network. When $r=\varnothing$, the risk latent is set to all zeros.


\subsection{Risk Criterion}
\label{Sec:Risk}

We then introduce how the risk level $r$ is defined. Generally, we can use a series of surrogate safety measures~\cite{wang2021review} to evaluate traffic safety, and this work utilizes PET as a specific metric. Precisely, PET measures the time difference between the moment when one vehicle leaves a conflict area and the moment when another vehicle
enters the same area~\cite{peesapati2018can}. In this work, given a joint trajectory $\boldsymbol{\tau}$, we use the minimum PET value across all trajectory pairs to quantify the risk $r$ for the entire environment. To effectively calculate this value, we adopt the following method. The spatial layout of the traffic scenario is rasterized into uniformly spaced grids, and a grid cell is marked as occupied by a vehicle when its shape overlaps with that cell. Then, by computing the time difference between different occupancy events, we determine the PET value for that grid cell. The minimum PET across all grid cells is then taken as the PET for the entire scenario. 

Previous data observations~\cite{peesapati2018can} have revealed that when the PET value is less than $1\,\mathrm{s}$, the crash risk is significantly correlated with the PET. Particularly, an exponential form of relationship can be established. Accordingly, we design the following heuristic criterion to measure the risk value:
\begin{equation}
\label{Eq:risk}
    r =  \exp\left({-k\cdot\max\left(0, \frac{\mathrm{PET}}{T}-\sigma\right)}\right),
\end{equation}
where we have the augment coefficient $k=5$, the bias $\sigma=0.05$, and the normalized limit $T=3.2$ in our experiments. Under this design, we have the risk value is above $0.3$ when $\mathrm{PET} < 1\,\mathrm{s}$, and thus in our experiments, we focus on the risk in the range of $[0.3,1]$. 

\subsection{Tokenized Dynamics Check}

\begin{figure}[t]
	\vspace{1mm}
	\centering
        \subfigure[Dynamics check]{
        \includegraphics[width=0.38\textwidth]{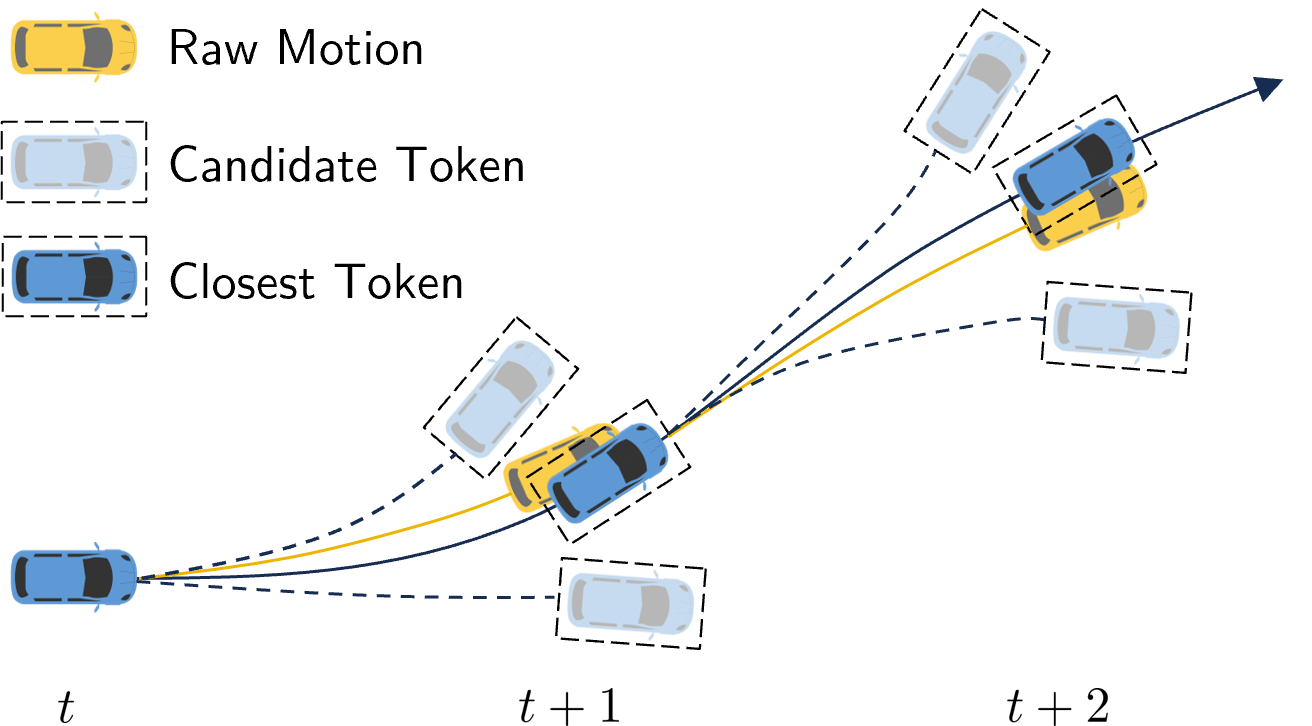}
        }
        \subfigure[Motion token vocabulary]{
        \includegraphics[width=0.44\textwidth]{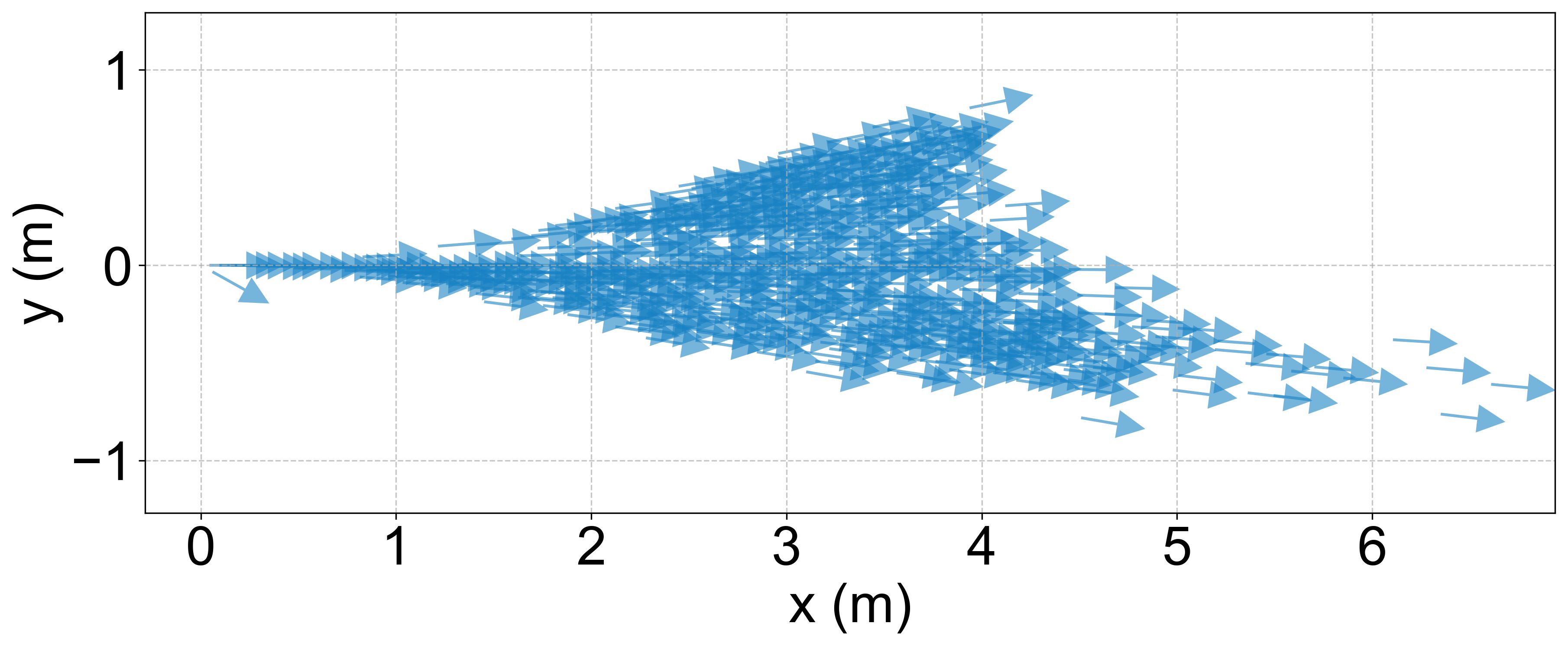}}
	\vspace{-2mm}
	\caption{Tokenized dynamics check. (a) Schematic for dynamics check. This module refines the raw trajectory by finding the closet motion token from a set of motion vocabulary. (b) Visualization of motion token vocabulary constructed from the rounD dataset~\cite{krajewski2020round}. The vocabulary contains $1024$ tokens, and here for better visualization, we show $512$ tokens.}
	\label{Fig:DynamicsCheck}
\end{figure}

For diffusion over states, it is essential to incorporate a dynamics check module to ensure that the generated trajectories satisfy the requirement of physical plausibility. Typical approaches enforce dynamics constraints by introducing a guidance function for trajectory optimization during the sampling process, which significantly increases computational burden due to the need for gradient calculation. 

For efficient dynamics check, we design a tokenized dynamics check module in this paper, motivated by recent GPT-style trajectory generation or prediction techniques~\cite{wu2024smart,philion2023trajeglish,shi2022motion}, which have demonstrated strong performance on discrete trajectory tasks, particularly in small datasets. As shown in Fig.~\ref{Fig:DynamicsCheck}(a), the core idea is to constrain trajectory generation within a set of valid state transitions, referred to as motion tokens. Specifically, we define a motion vocabulary $V=\{\boldsymbol{a}\}$, where each token $\boldsymbol{a}$ represents a state transition from current state $\boldsymbol{s}$ to next state $\boldsymbol{s}'$. This vocabulary is constructed by applying a k-disks clustering algorithm~\cite{philion2023trajeglish} to all motion transitions observed in the dataset; see Fig.~\ref{Fig:DynamicsCheck}(b) for illustration of the constructed motion vocabulary.

\begin{figure*}[!t]
	\vspace{1mm}
	\centering
     \subfigure[Scenario 1 (low risk)]
	{\includegraphics[width=0.23\textwidth]{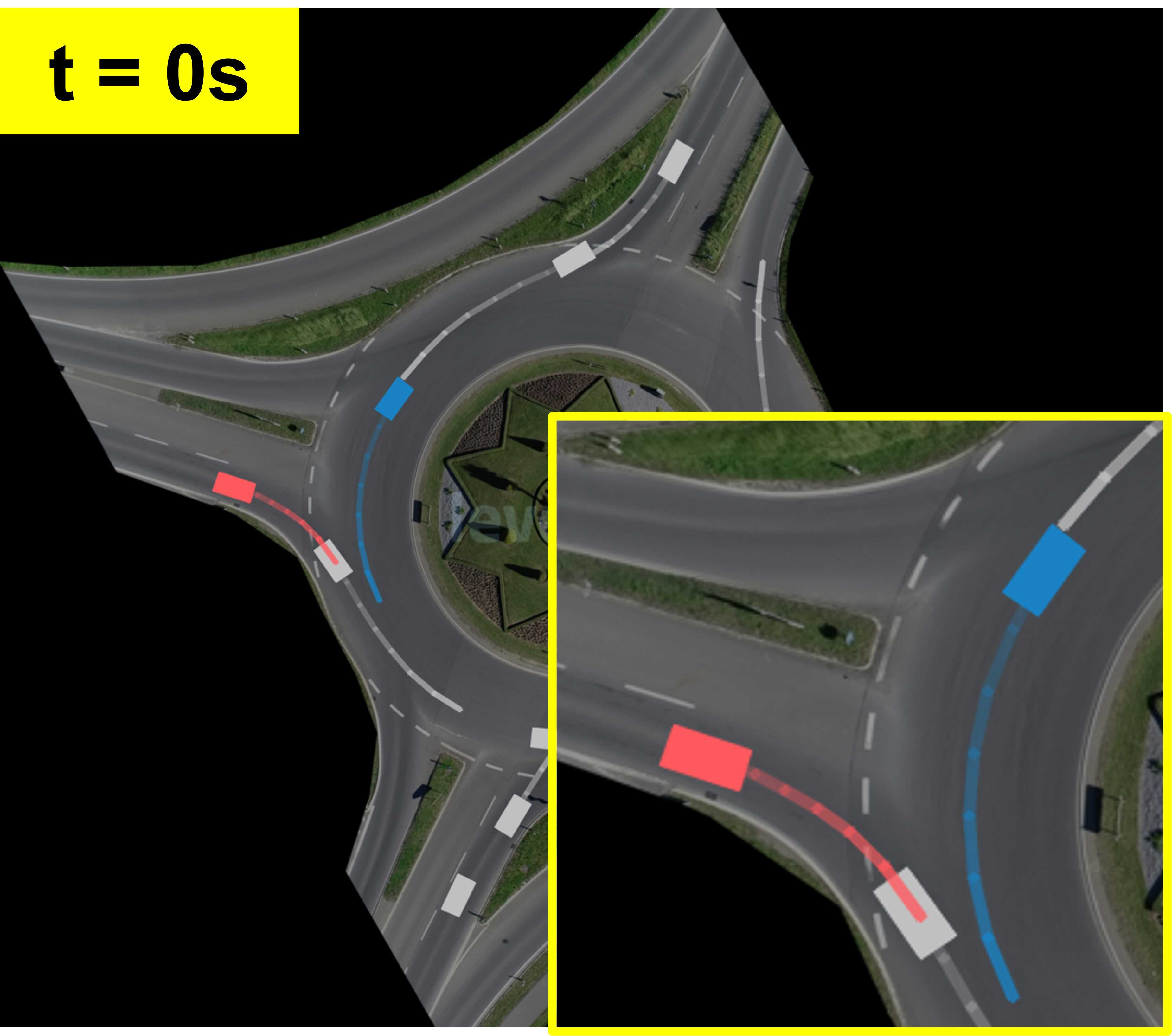}
    \includegraphics[width=0.23\textwidth]{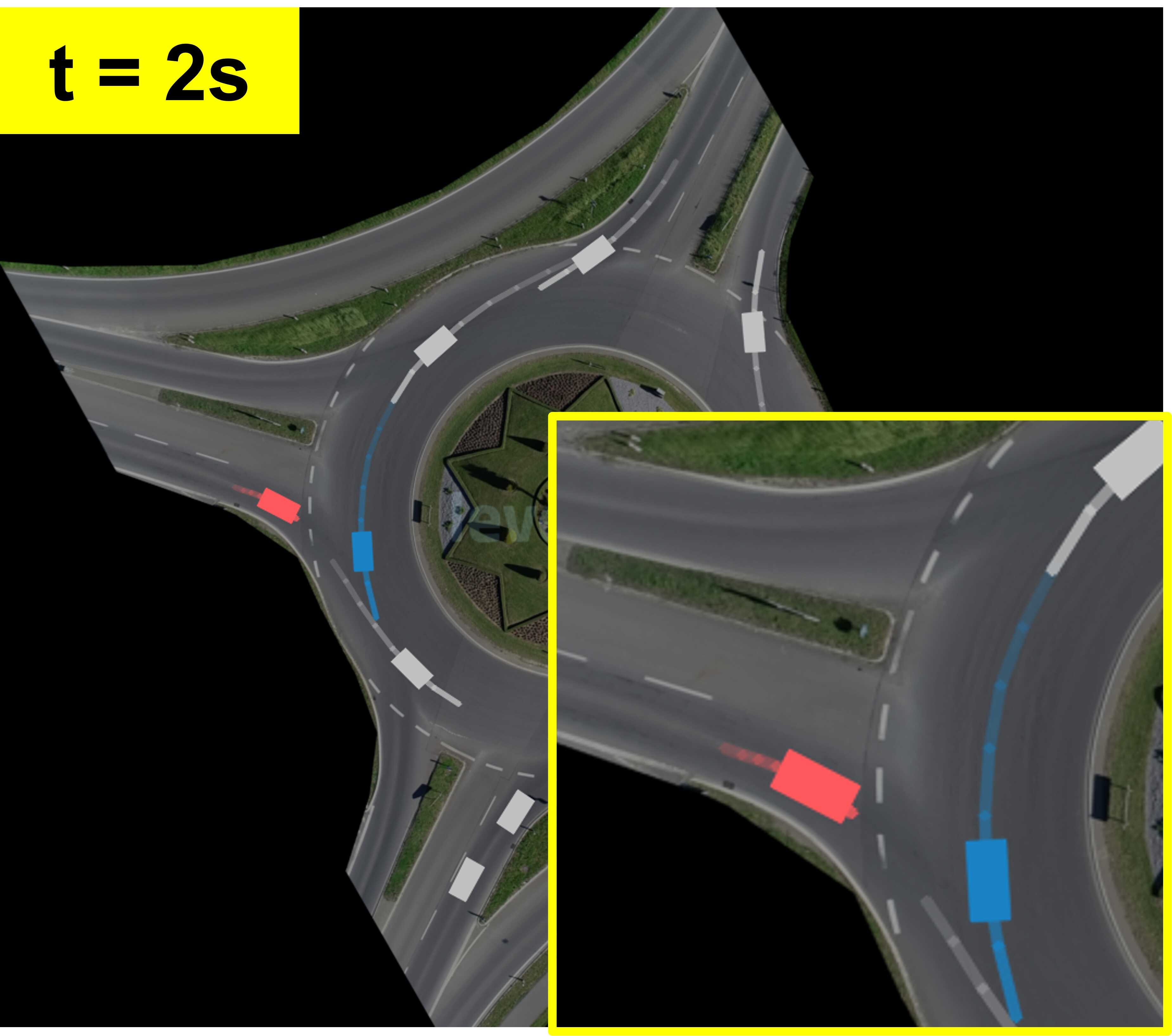}
	}
    \hspace{1mm}
	\subfigure[Scenario 1 (high risk)]
	{\includegraphics[width=0.23\textwidth]{Figures/edited-show_future_0_data-92-desiredrisk-0.3-sample-4-denoising-0-actualPET-1.0-compressed.jpg}
    \includegraphics[width=0.23\textwidth]{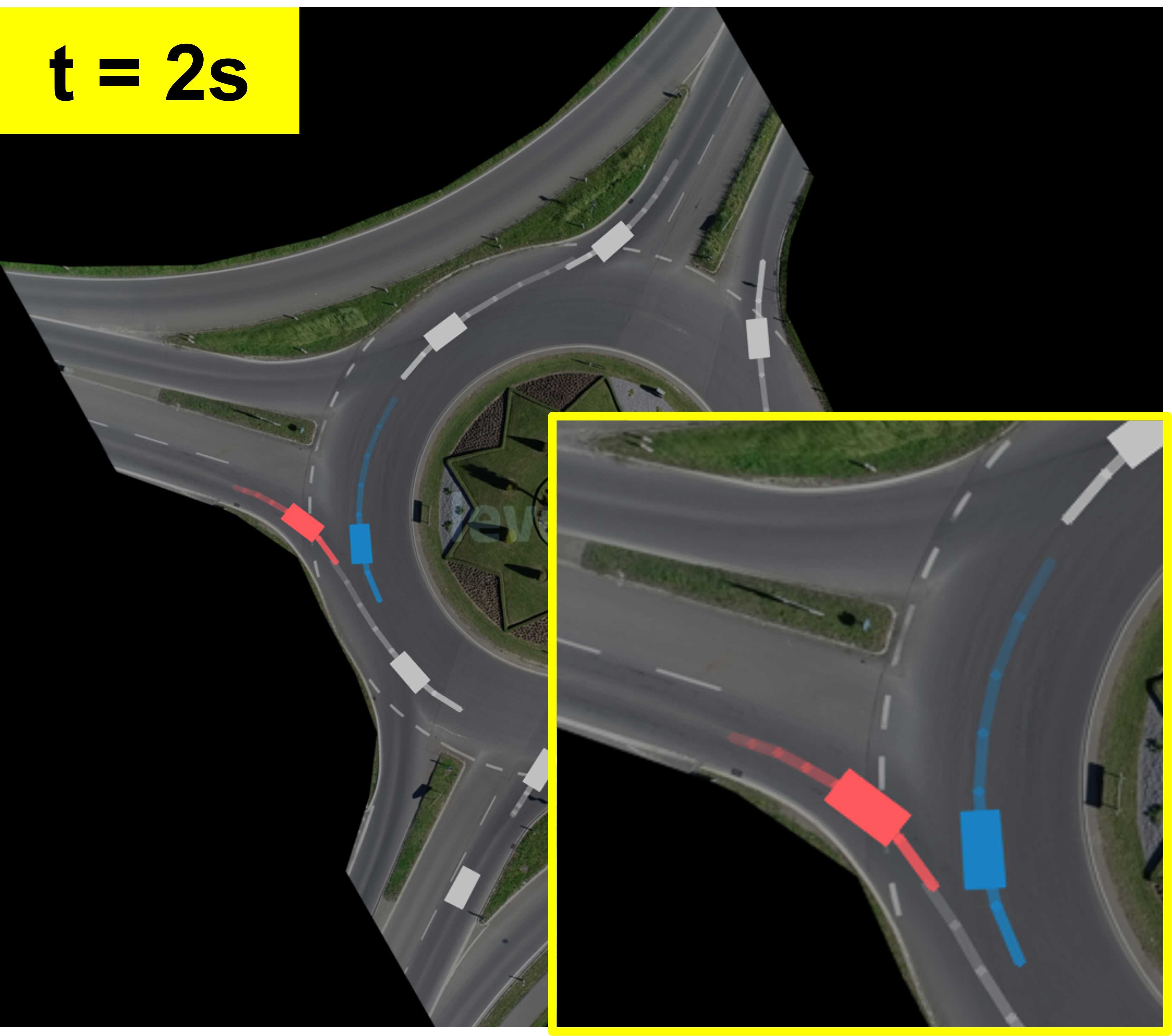}
	}
	\subfigure[Scenario 2 (low risk)]
	{\includegraphics[width=0.23\textwidth]{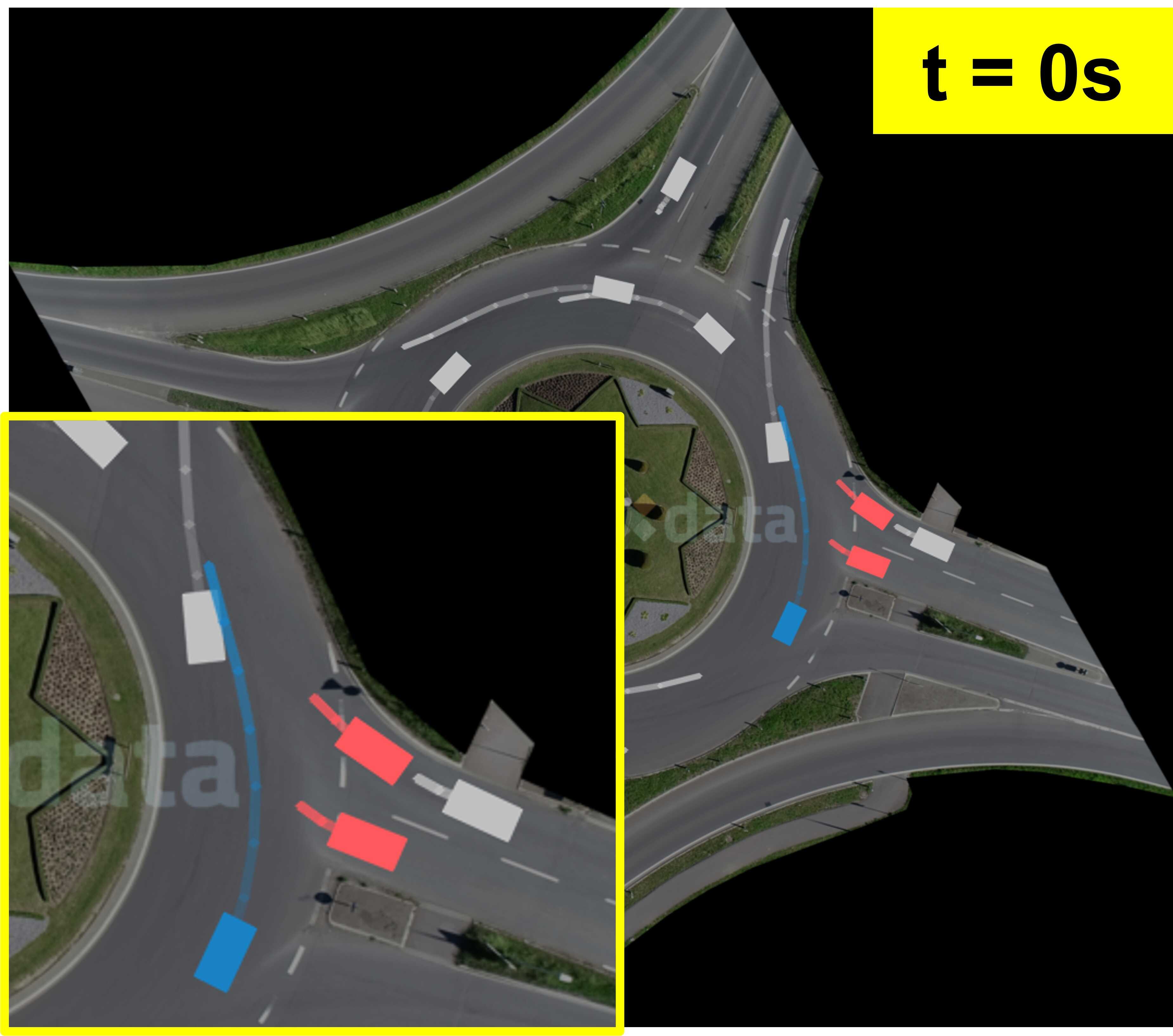}
    \includegraphics[width=0.23\textwidth]{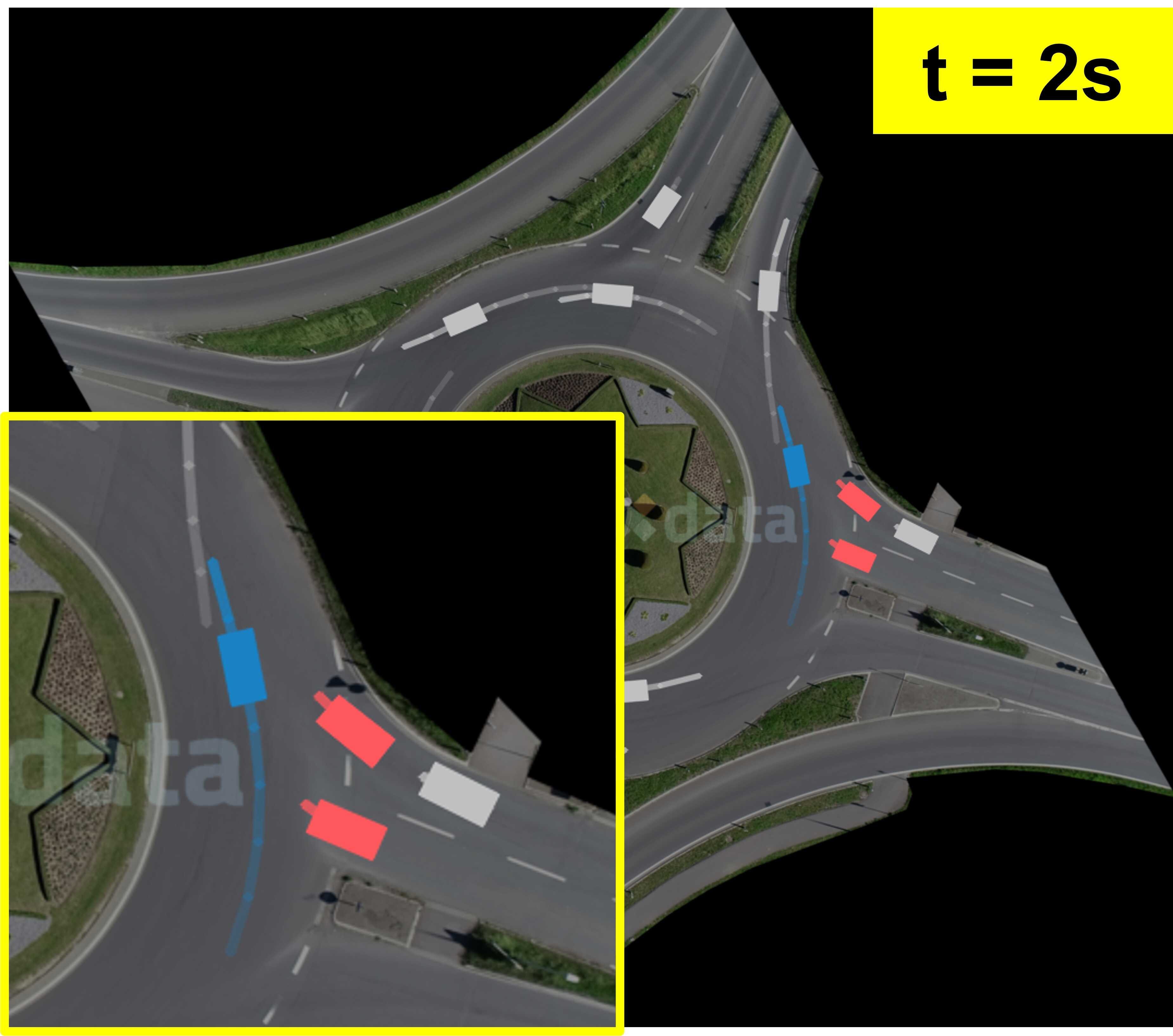}
	}
    \hspace{1mm}
	\subfigure[Scenario 2 (high risk)]
	{\includegraphics[width=0.23\textwidth]{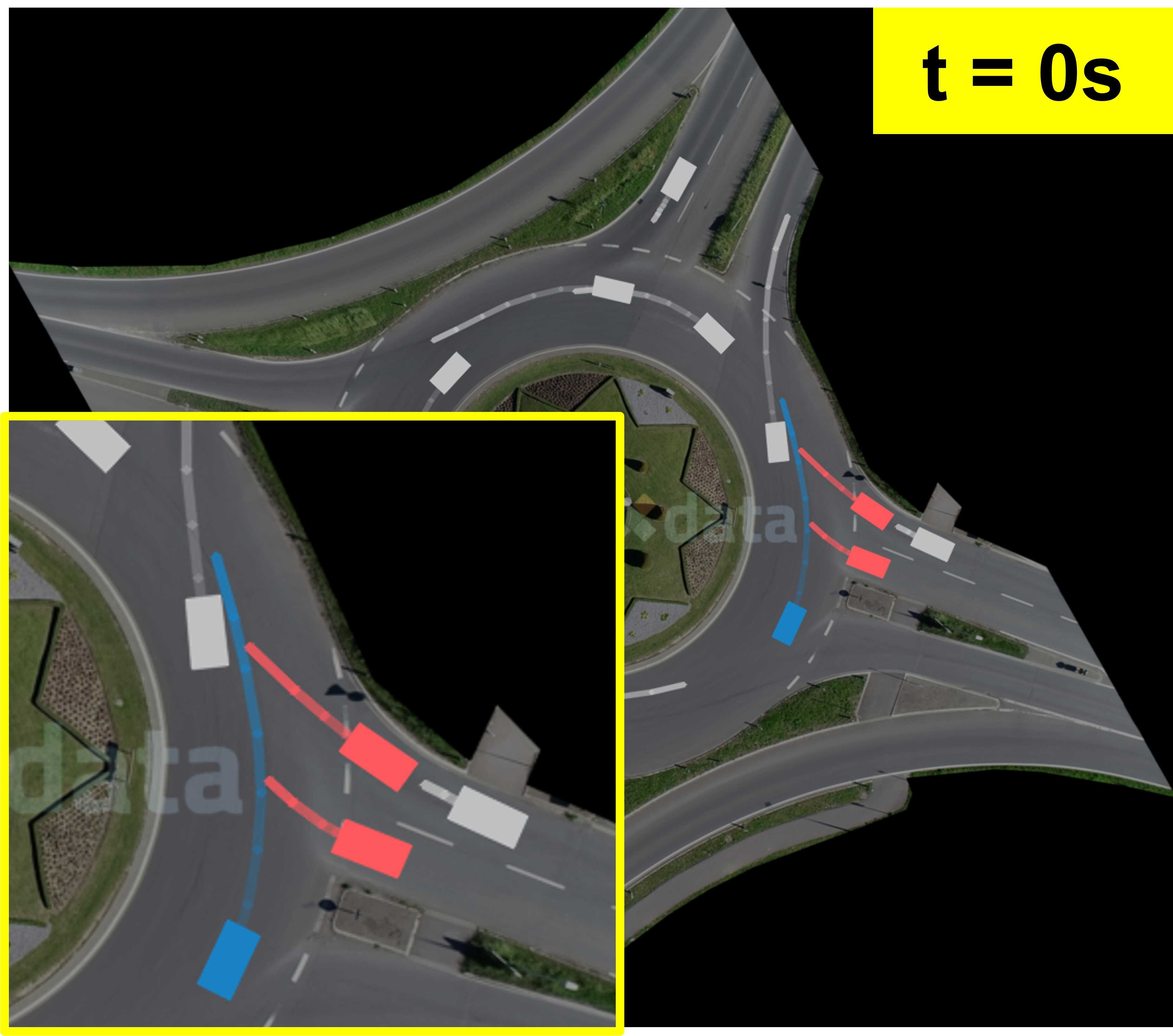}
    {\includegraphics[width=0.23\textwidth]{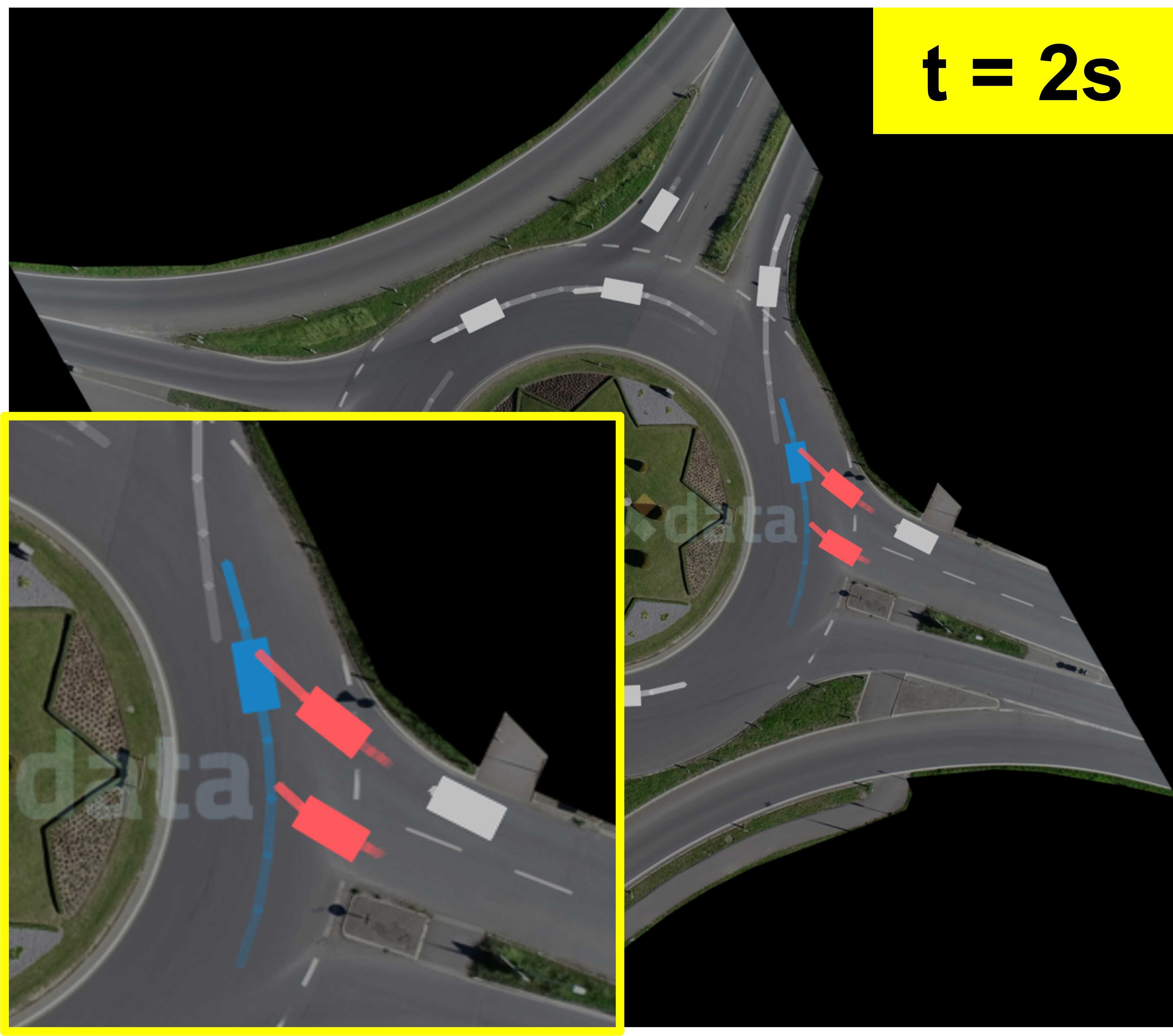}}
	}
	\caption{Snapshots of two scenarios under different risk levels. For each scenario, the left and right panels represent vehicle states at $t=0\,\mathrm{s}$ and $t=2\,\mathrm{s}$, respectively. In each case, RADE is used once to generate the future joint trajectories, visualized as colored lines. Vehicles exhibiting more risk-inducing behaviors are highlighted in red. The comparisons between low-risk (left columns) and high-risk (right columns) setups illustrate how RADE adjusts agent behavior based on the specified risk level.}
	\label{Fig:casestudy}
\end{figure*}

Then, given current state $\boldsymbol{s}$ and a proposed next state $\boldsymbol{s}'$ from the generative neural network, we calculate the corresponding proposed motion $\boldsymbol{a}'$ and search for the most similar motion token in the vocabulary $V$. The selected motion token  $\boldsymbol{a}^*$ is determined by
\begin{equation}
    \boldsymbol{a}^* = \arg\min_{\boldsymbol{a} \in V} d( \boldsymbol{a} - \boldsymbol{a}' ),
\end{equation}
where $d$ measures the average corner distance between a box of length $3.6\,\mathrm{m}$ and width $1.8\,\mathrm{m}$ with respect to the action. 
Using current state $\boldsymbol{s}$ and the selected motion $\boldsymbol{a}^*$, we obtain the corresponding valid next state $\boldsymbol{s}^*$. This valid next state then serves as the reference to iteratively compute the next time step's state. The entire trajectory generated by the diffusion network is passed through this auto-regressive correction process to improve the dynamics feasibility. With the integration of this module, the closed-loop simulation procedure is summarized in Algorithm~\ref{Alg:Simulation}, where only the first future state from the generated trajectory is used for updating at each time step.

\begin{algorithm}[t]
\caption{Closed-Loop Simulation}
\label{Alg:Simulation}
\begin{algorithmic}[1]
\State \textbf{Input:} Noise model $\epsilon_{\theta}$, guidance scale $\omega$, risk $r$
\State Initialize current state $\hat{\boldsymbol{\tau}}; \ t \gets 0$ \hfill 
\While{not done}
    \State Initialize $\boldsymbol{\tau}^K \sim \mathcal{N}(0, \alpha I)$
    \For{$k = K \dots 1$}
        \State $\boldsymbol{\tau}^k[0] \gets \hat{\boldsymbol{\tau}}$ \hfill 
        \State $\hat{\epsilon} \gets \epsilon_{\theta}(\boldsymbol{\tau}^k, k) + \omega\left(\epsilon_{\theta}(\boldsymbol{\tau}^k, r, k) - \epsilon_{\theta}(\boldsymbol{\tau}^k, k)\right)$ \hfill 
        \State $(\boldsymbol{\mu}_{k-1}, \Sigma_{k-1}) \gets \text{Denoise}(\boldsymbol{\tau}^k, \hat{\epsilon})$
        \State $\boldsymbol{\tau}^{k-1} \sim \mathcal{N}(\boldsymbol{\mu}_{k-1}, \alpha \Sigma_{k-1})$
    \EndFor
    \State Tokenized dynamics check for $\boldsymbol{\tau}^0$
    \State Update current state $\hat{\boldsymbol{\tau}} \gets \boldsymbol{\tau}^0[1]$; $t \gets t + 1$
\EndWhile
\end{algorithmic}
\end{algorithm}


\section{Experiments}
\label{Sec:5}

We now present the experiments of RADE on the real-world dataset. Our results show that RADE can generate trajectories given varying risk levels, while preserving statistical realism.

\subsection{Experiment Settings}

We conduct our experiments using the rounD dataset\cite{krajewski2020round}, which records vehicle trajectories at a two-lane roundabout in Neuweiler, Aachen, Germany. This environment features dense interactions and complex negotiation behaviors, making it well-suited for evaluating risk-adjustable traffic simulations. Both the dataset and our simulation operate at a time resolution of $0.4\,\mathrm{s}$, and the time horizon for generated trajectory is $3.2\,\mathrm{s}$ ($8$ steps). As noted in our risk formulation~\eqref{Eq:risk}, meaningful risk is observed when the Post-Encroachment Time (PET) falls below $1$ second, motivating our focus on a risk index range of $r \in [0.3,1]$. Unless otherwise specified, we refer to low-risk setting as $r=0.3$ and high-risk setting as $r=1.0$ in this section.

\subsection{Case Study}

We begin by presenting representative examples to illustrate RADE’s ability to generate vehicle trajectories under varying risk levels. As shown in Fig.~\ref{Fig:casestudy}, each scenario has the same initial state at 
$t=0\,\mathrm{s}$, and RADE is used to generate the trajectories for all vehicles in a single forward pass (no autoregressive generation as in closed-loop simulation). In the low-risk setting, vehicles exhibit cautious behavior, and no apparent risk is observed throughout the scenario. In contrast, under a high-risk setting, RADE naturally induces more aggressive behaviors in certain vehicles (highlighted in red), such as attempting to enter the roundabout despite another vehicle approaching at close distance inside the roundabout. This behavior significantly increasing the likelihood of a collision. Importantly, these high-risk behaviors are not explicitly assigned to any vehicle. Instead, they emerge from the joint trajectory generation across all vehicles, demonstrating RADE’s ability to generate plausible  behaviors for the entire scenario given a desired risk level.

\begin{figure}[t]
	\vspace{1mm}
	\centering
    \subfigure[Distance]
	{\includegraphics[width=0.22\textwidth]{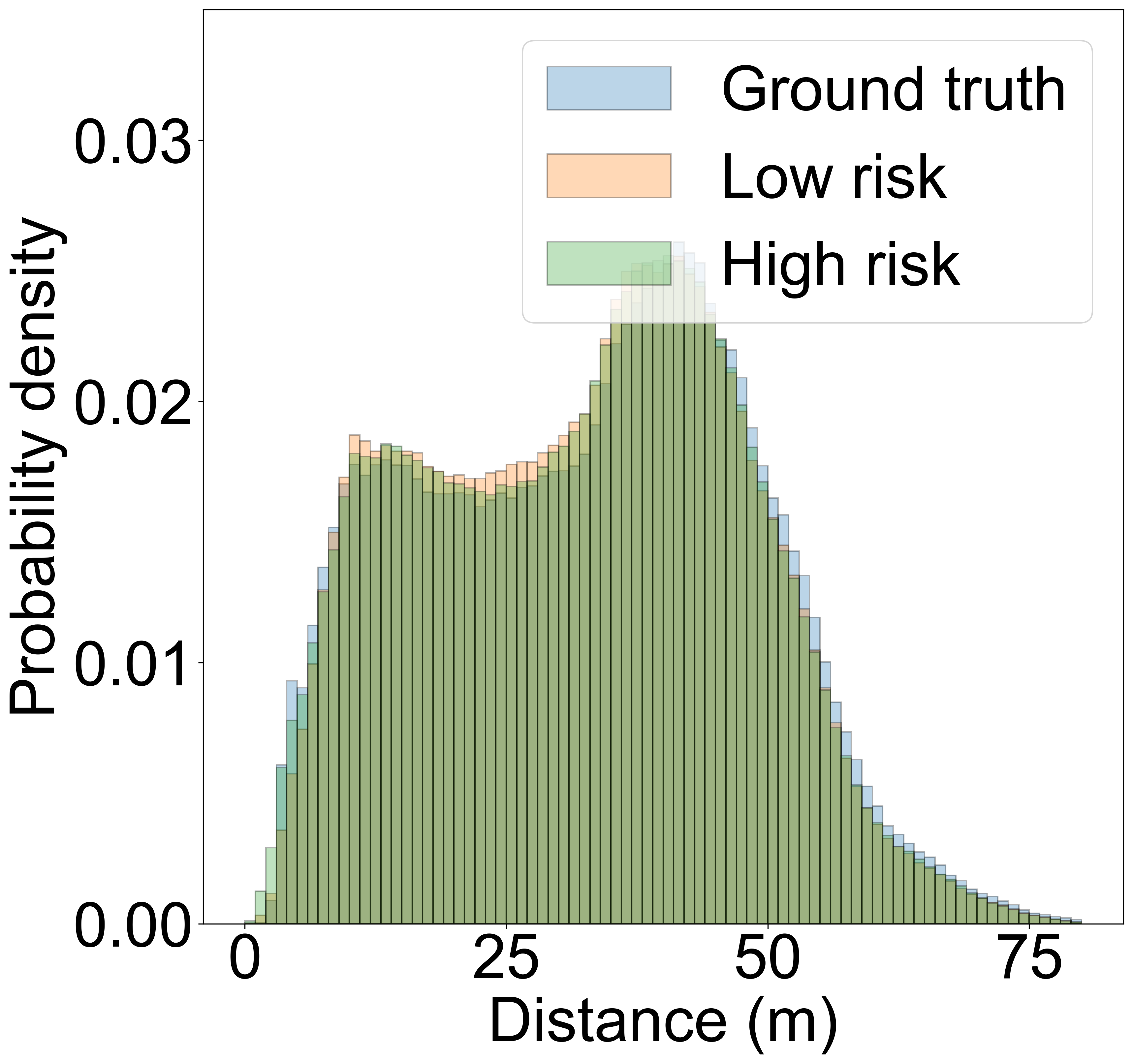}
}
	\subfigure[Speed]
	{\includegraphics[width=0.22\textwidth]{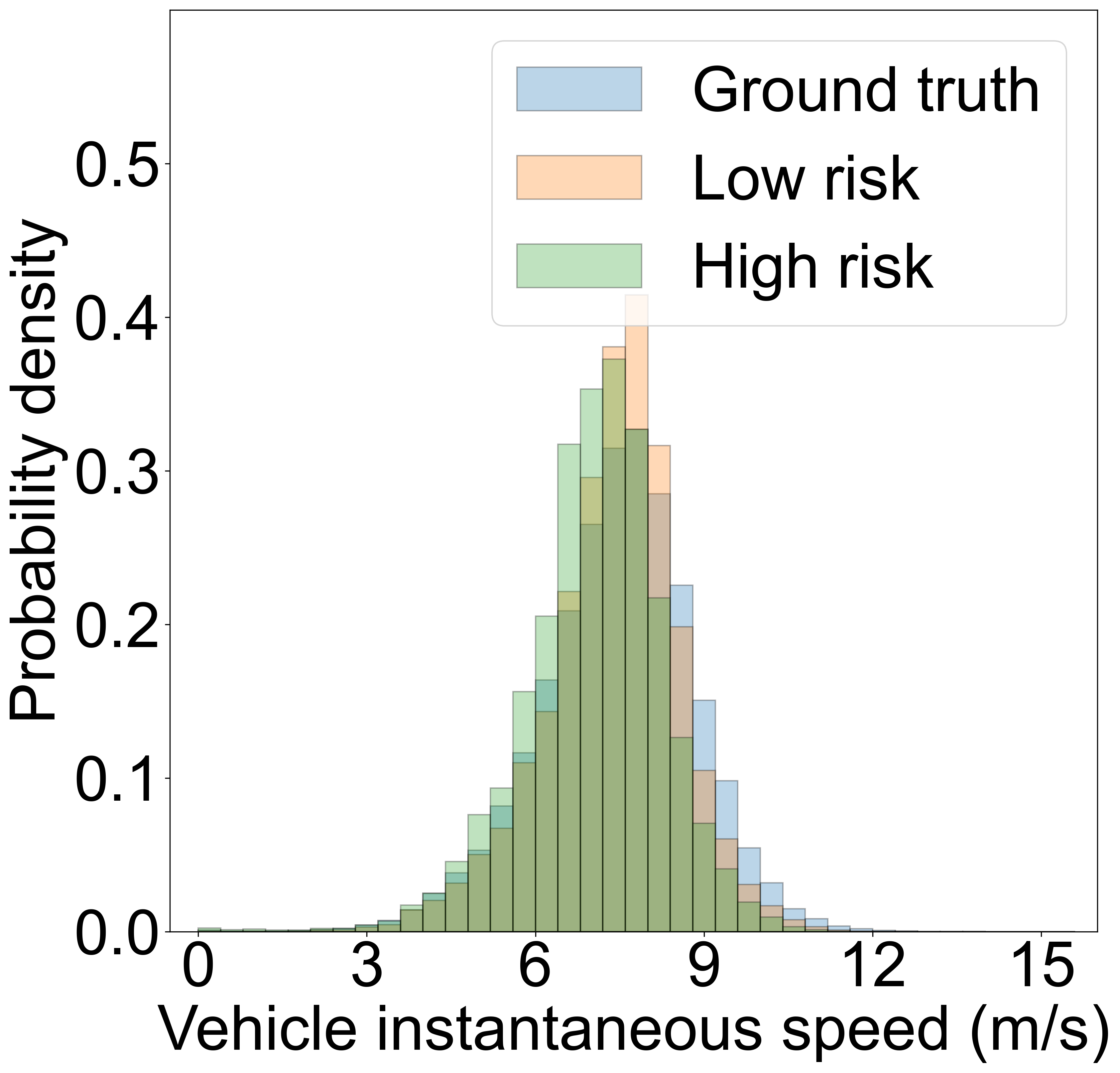}
}
	\subfigure[Yielding distance]
	{\includegraphics[width=0.22\textwidth]{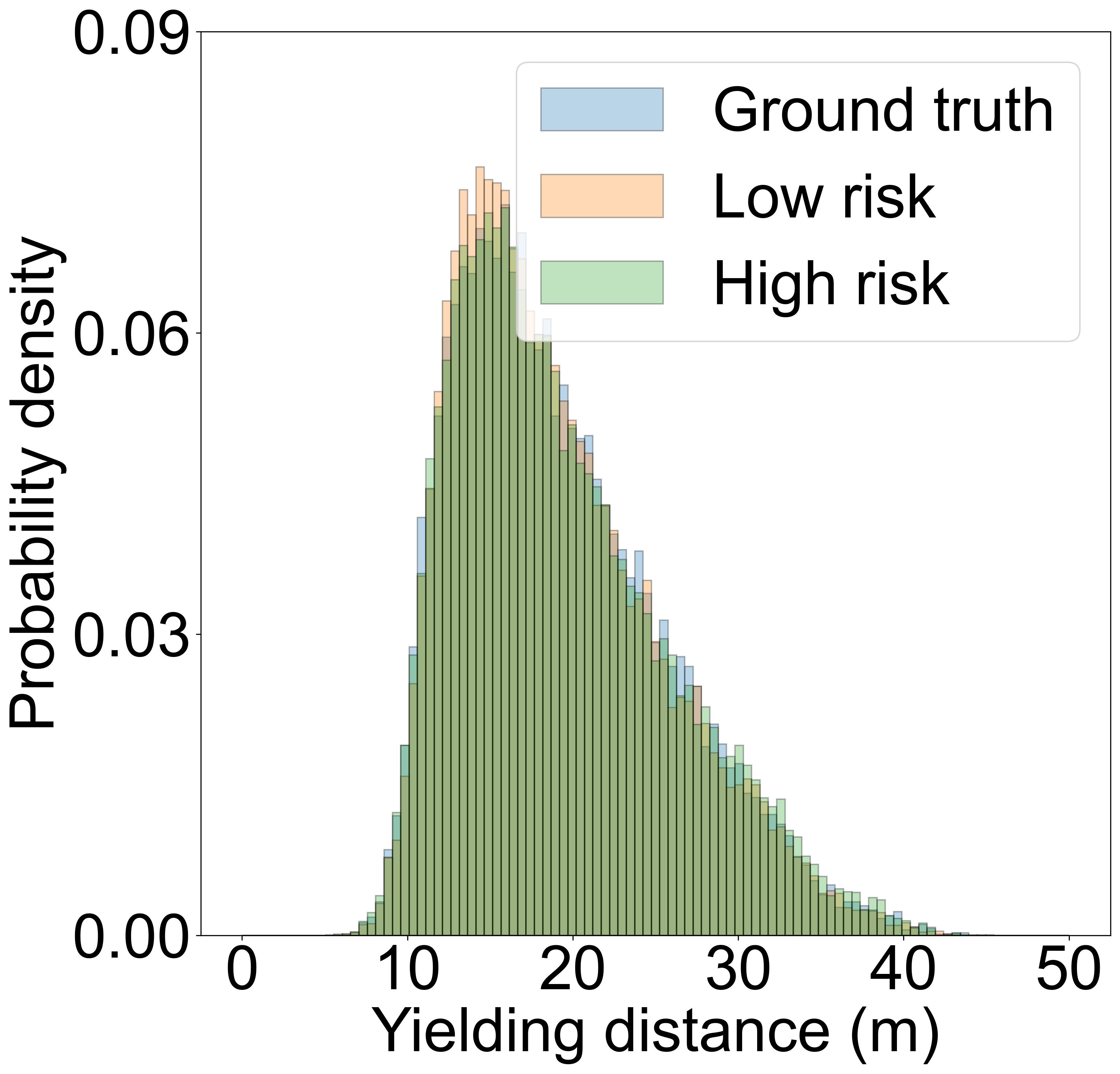}
}
	\subfigure[Yielding speed]
	{\includegraphics[width=0.22\textwidth]{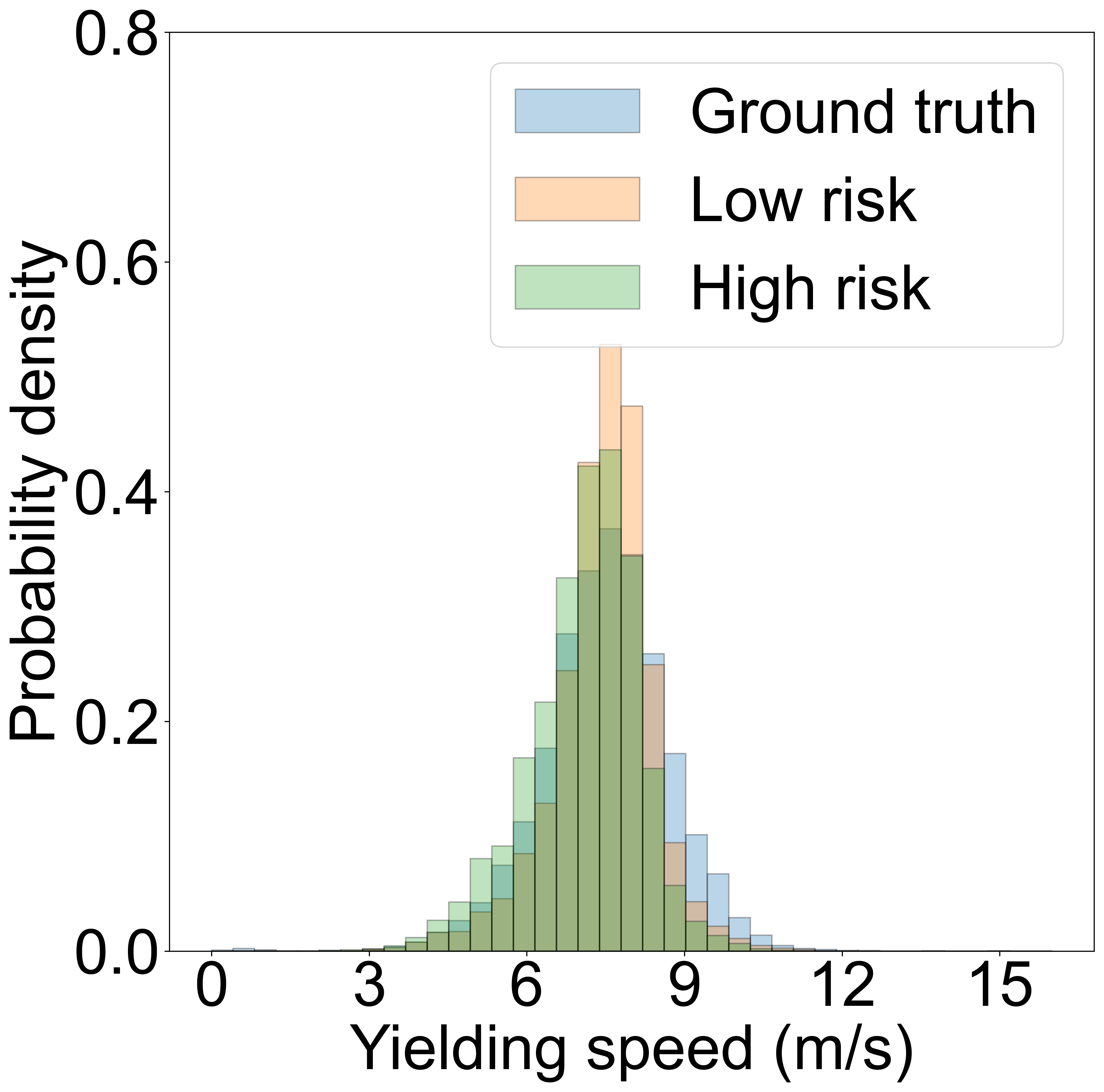}
}
	\caption{Statistical realism at different risk levels. Each metric is calculated by aggregating results from $10$ random-seeded experiments. The distance distribution is closely aligned with real-world data at different risk levels, and the speed distribution exhibits a similar overall trend.}
	\label{Fig:StatisticalRealism}
\end{figure}

\subsection{Closed-Loop Simulations}

We then evaluate the performance of RADE in a closed-loop setup. Precisely, the proposed RADE simulator is initialized using a randomly sampled $2$-second trajectory clip, during which all vehicles follow their ground-truth trajectories. After initialization, the behavior of all vehicles is generated by RADE following Algorithm~\ref{Alg:Simulation}. At each simulation step, new vehicles are introduced at entry lanes based on a Poisson arrival process, which is calibrated from the dataset. Vehicles that reach designated exit zones are removed from the scenario. We conduct $10$ random-seeded experiments and for each experiment, we run $100$ random simulation episodes, each lasting $30$ seconds, across varying risk levels $r\in\{0.3,0.4,\ldots,1.0\}$.  

\vspace{0.2em}
\noindent\textbf{Statistical Realism.} Since high-fidelity modeling of driving behaviors is essential for realistically simulating safety-critical scenarios, we begin by evaluating the statistical realism of RADE-generated traffic across varying risk levels. We focus on two key microscopic metrics: inter-vehicle distance and instantaneous vehicle speed. As shown in Fig.~\ref{Fig:StatisticalRealism}(a), RADE achieves a highly accurate match with the real-world distribution of inter-vehicle distances, indicating that it effectively preserves spatial interaction patterns among vehicles. This strong alignment holds consistently across different risk levels, demonstrating that increased risk does not compromise the naturalistic property of inter-vehicle spacing. While the speed distribution (Fig.~\ref{Fig:StatisticalRealism}(b)) shows some slight deviation from the ground truth, it still captures the overall trend and variation in driving speeds.

To further evaluate interaction fidelity, we examine the statistical realism of the yielding behavior at roundabout entrance areas, where complex interactions frequently occur. Specifically, we focus on the distributions of yielding distance and yielding speed. The yielding distance is defined as the distance between the yielding vehicle at the entrance and the nearest conflicting vehicle in the roundabout, and the yielding speed is defined as the speed of the nearest conflicting vehicle. The distribution results are shown in Fig.~\ref{Fig:StatisticalRealism}(c)(d). It can be clearly observed that RADE produces a yielding behavior distribution that closely matches the real-world data. These results indicate RADE captures realistic driving behaviors and well preserves the fundamental requirement of statistical realism, despite different setups of the desired risk levels.

\begin{figure}[t]
	\vspace{1mm}
	\centering
	\subfigure[Consistent traffic volume]
	{\includegraphics[width=0.22\textwidth]{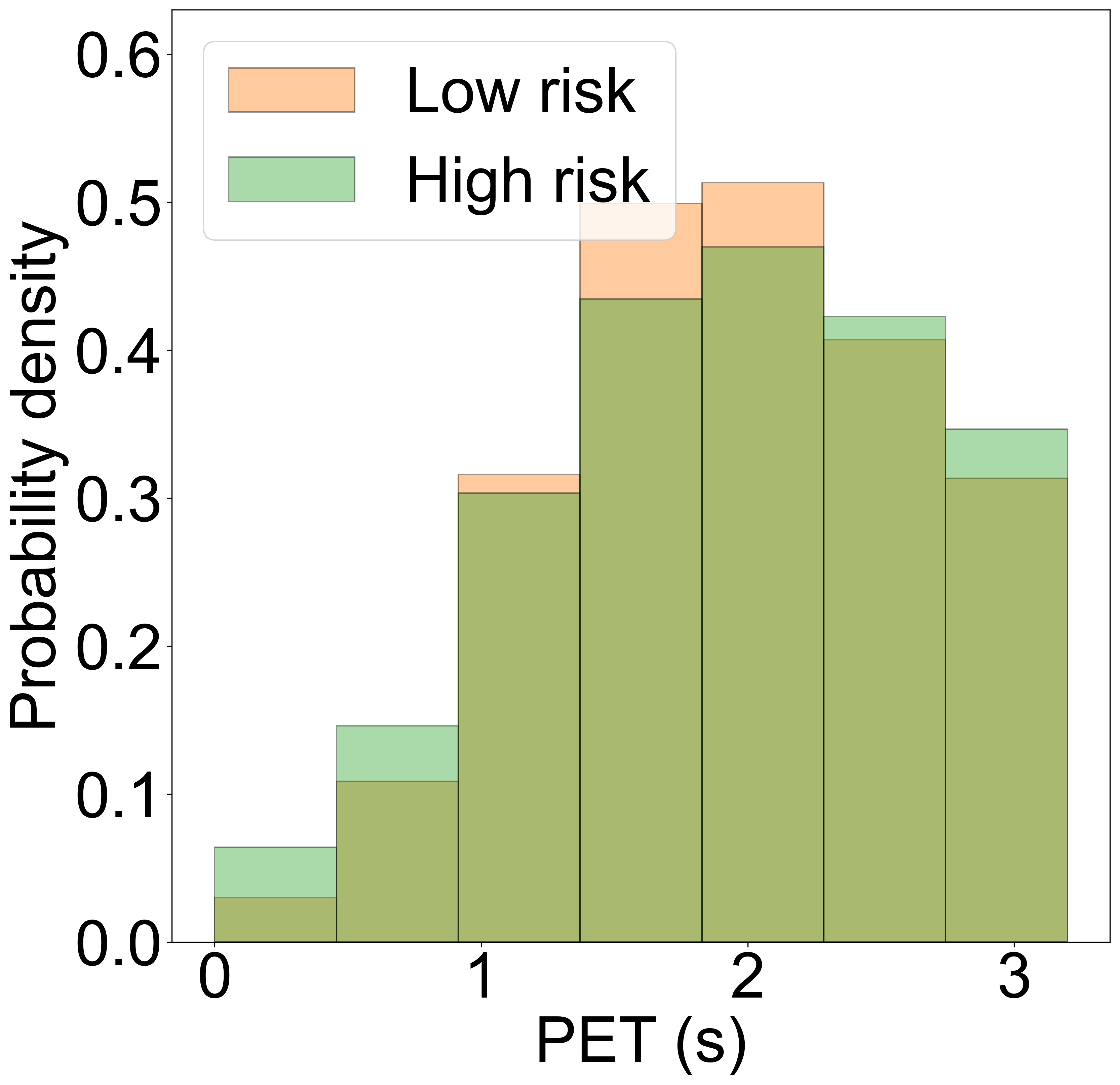}
}
	\subfigure[Varying traffic volume]
	{\includegraphics[width=0.22\textwidth]{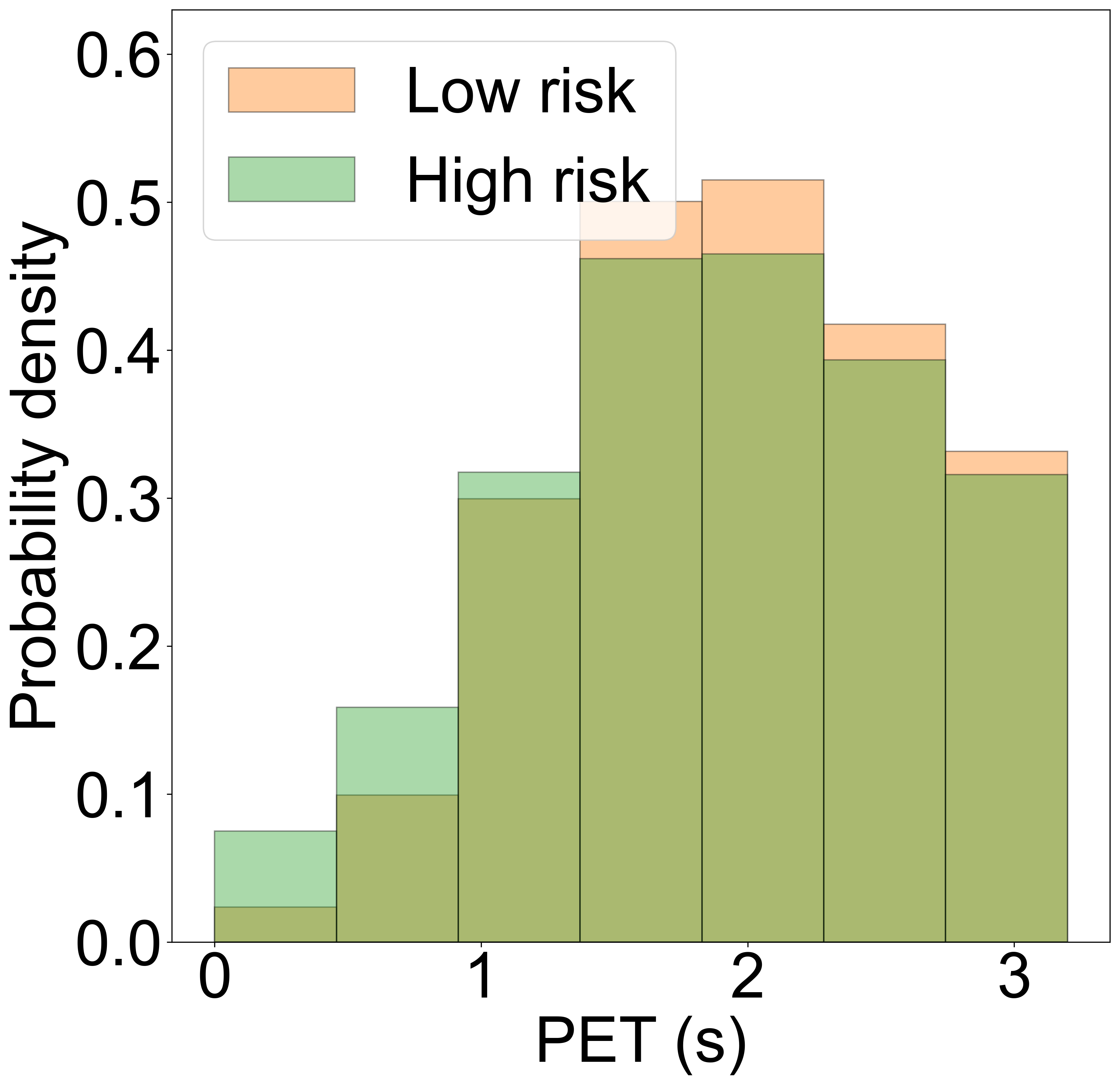}
}
	\caption{Comparison of PET distribution at different risk levels. Each metric is calculated by aggregating results from $10$ random-seeded experiments. The high-risk setting exhibits a greater concentration of PET values below $1\,\mathrm{s}$, indicating a higher likelihood of safety-critical interactions.  }
	\label{Fig:PETDistribution}
\end{figure}

\vspace{0.2em}
\noindent\textbf{Adjustable Risk.} We proceed to evaluate whether the actual traffic risk has been successfully controlled in RADE by analyzing the PET distribution and the crash rate under varying risk levels. As shown in Fig.~\ref{Fig:PETDistribution}(a), in the high-risk setting, PET values are more densely concentrated in the region below $1\,\mathrm{s}$, which, as previously mentioned, is a typical threshold for identifying those interactions strongly associated with collision likelihood. Particularly, at the most critical region of $\mathrm{PET}<0.4$, which captures near-miss events, the probability density is increased by $113\%$ from $0.030$ at low risk setup to $0.064$ at high risk setup. This result indicates that RADE effectively generates scenarios with closer time gaps between interacting vehicles, thereby increasing the possibility for safety-critical events.

\begin{figure}[t]
	\vspace{1mm}
	\centering
	{\includegraphics[width=0.45\textwidth]{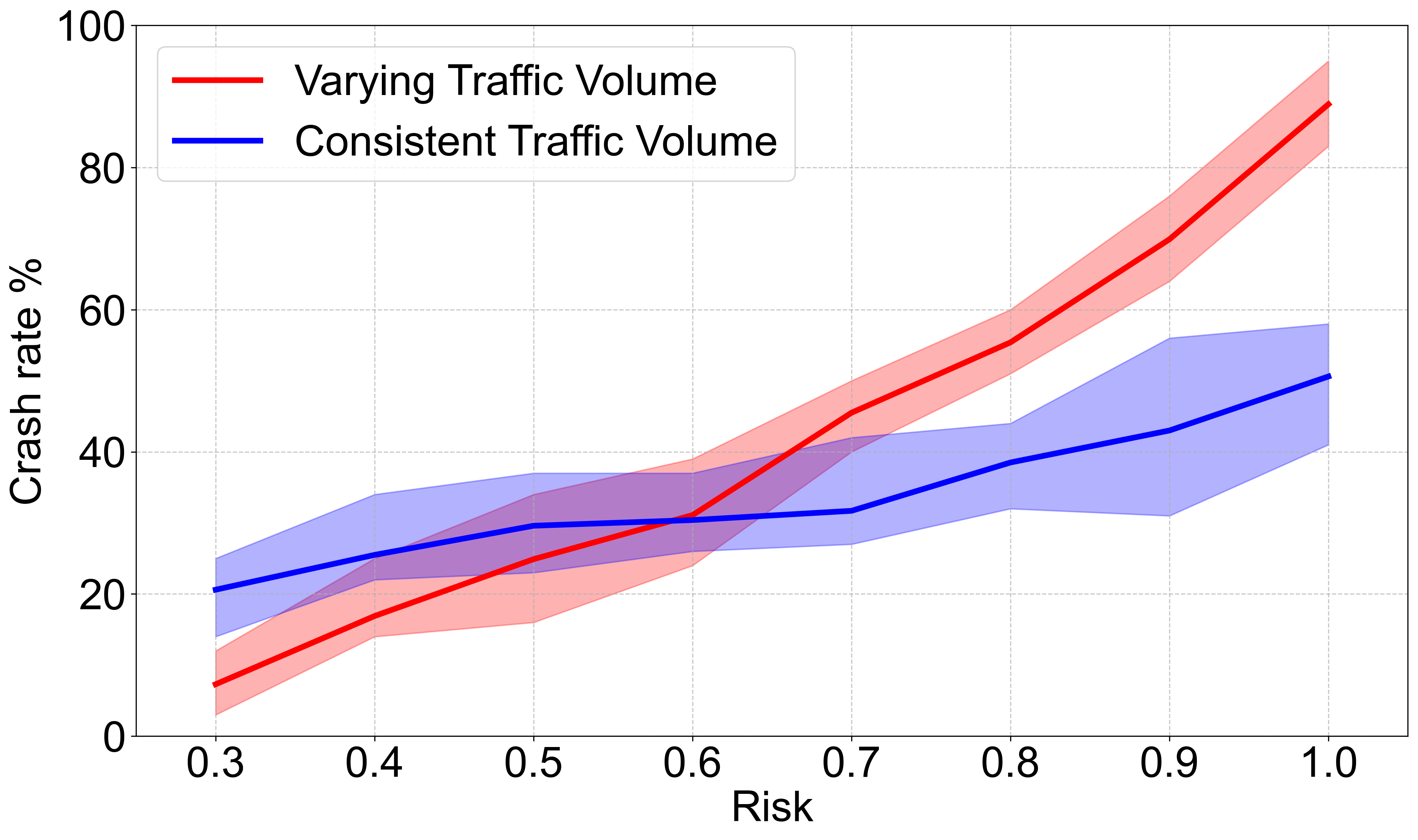}
	}
	\caption{Crash rate across varying risk levels under two traffic volume settings. The solid lines represent the average crash rate over 10 random-seeded experiments, while the shaded regions indicate the range between the minimum and maximum value. The red curve corresponds to varying traffic volume, linearly scaled with risk, and the blue curve corresponds to a consistent traffic volume setting. RADE demonstrates effective control over crash rates by adjusting the input risk level, with a stronger effect under varying traffic volume. }
	\label{Fig:CrashRate}
\end{figure}

Further, in our setup, a simulation episode is labeled as a crash episode if one collision occurs between any pair of vehicles. The crash rate is then defined as the proportion of crash episodes over the total number of simulation episodes. As shown in Fig.~\ref{Fig:CrashRate}, the crash rate consistently increases as the desired risk level grows up, with the mean value rising from $20.6\%$ to $50.6\%$. This confirms that RADE effectively controls the emergence of risk through its risk-conditioning mechanism. It is worth noting that these crashes are not manually injected through explicit manipulation or adversarial activation, as is common in most adversarial testing frameworks. Instead, they emerge naturally from the jointly generated behaviors of multiple vehicles in response to the conditioned risk level. This enables RADE to create realistic and diverse safety-critical scenarios in a scalable manner. This is essential for efficient and accurate AV safety testing, as it allows AVs to be exposed to a broad range of plausible yet challenging traffic conditions, without relying on predefined adversaries or manual interventions.

As suggested in the literature~\cite{paul2020post}, crash likelihood also has a close relationship with traffic volume. We then conduct an additional experiment where the traffic volume is linearly scaled with the desired risk level. Specifically, the lowest-risk setting uses 0.5× the baseline of arrival rates, while the highest-risk setting uses 1.5×, with intermediate values interpolated linearly. The initial vehicle number is adjusted accordingly. As illustrated in Fig.~\ref{Fig:PETDistribution}(b) and Fig.~\ref{Fig:CrashRate}, this setup amplifies the distinction in PET distributions across risk levels within the PET range below $1\,\mathrm{s}$. Particularly, at the most critical region of $\mathrm{PET}<0.4$, the probability density rises from $0.024$ at low risk to $0.075$ at high risk, indicating an over $212\%$ likelihood increase of near-miss events. Moreover, the crash rate is controlled across a wider range, with the average value increasing from $7.3\%$ to $88.9\%$ as the risk level grows up, highlighting RADE’s ability to effectively control risk exposure.

\begin{figure}[t]
	\vspace{1mm}
	\centering
    \subfigure[Angle crash]
	{\includegraphics[width=0.15\textwidth]{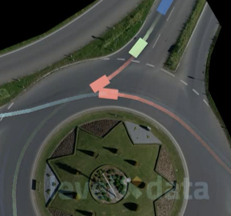}
    \includegraphics[width=0.15\textwidth]{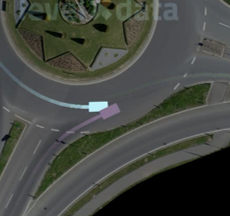}
    \includegraphics[width=0.15\textwidth]{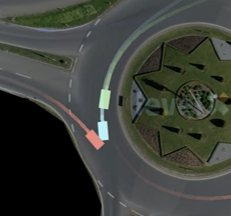}
}
	\subfigure[Sideswipe crash]
	{\includegraphics[width=0.15\textwidth]{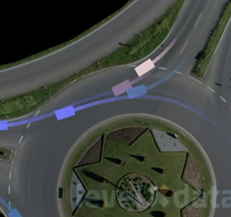}
    \includegraphics[width=0.15\textwidth]{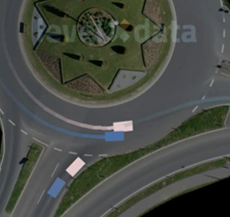}
    \includegraphics[width=0.15\textwidth]{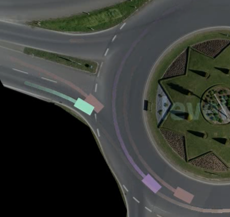}
}
	\caption{Examples of crashes generated by RADE. (a) Angle crashes, which typically occur when a vehicle enters the roundabout without yielding to cross-traffic. (b) Sideswipe crashes, resulting from vehicles failing to maintain proper lateral clearance while navigating in parallel. }
	\label{Fig:CrashExample}
\end{figure}

Finally, we visualize some examples of the most common crash types observed from our simulations at the high-risk setting. According to traffic incident records for roundabouts, angle crashes and sideswipe crashes are among the most common collision types, often resulting from failures of complex negotiation or gap maintenance during close interactions between vehicles. As shown in Fig.~\ref{Fig:CrashExample}, RADE is able to reproduce both angle and sideswipe crashes. Again, these crashes are not the result of manual adversarial injections or behavior modifications. Instead, they naturally emerge from the jointly generated trajectories, conditioned solely on the desired risk level.
These examples illustrate RADE’s capability to simulate realistic and diverse crash scenarios, which is crucial for comprehensive and scalable validation for AV safety.


\section{Conclusions and Discussions}
\label{Sec:6}

In this work, we propose the Risk-Adjustable Driving Environment (RADE), to generate statistically realistic and risk-adjustable multi-agent traffic scenarios. Conditioned a PET-based risk level, RADE leverages a multi-agent diffusion framework to generate joint future trajectories. A tokenized dynamics check module is introduced to ensure that the generated motions remain physically plausible. 
Extensive experiments on the real-world rounD dataset demonstrate that across varying risk levels, RADE well preserves key behavioral distributions and can generate safety-critical behaviors in a natural  way, without requiring manual adversarial design. These results validate RADE's capability in generating realistic and risk-adjustable traffic environments.

For future work, we aim to incorporate more context information such as maps and traffic signals. Another promising direction is to explore alternative risk measurements beyond PET, enabling more accurate characterization of safety-critical scenarios across diverse driving environments. Finally, we will apply RADE to localized regions around the autonomous vehicle to support targeted stress testing under varying levels of risk, achieving robust AV evaluation and training in complex real-world scenarios.








\bibliographystyle{IEEEtran}
\bibliography{IEEEabrv,mybibfile}

\end{document}